\documentclass[runningheads]{llncs}
\usepackage[T1]{fontenc}
\usepackage{amsmath,amssymb,amsfonts}
\usepackage{algorithmic}
\usepackage{graphicx}
\usepackage{textcomp}
\usepackage{xcolor}
\usepackage{subcaption}
\usepackage{hyperref}
\usepackage{url}
\usepackage{booktabs}
\usepackage{multirow}
\usepackage[normalem]{ulem}
\usepackage{wrapfig}

\newcommand\blfootnote[1]{%
  \begingroup
  \renewcommand\thefootnote{}\footnote{#1}%
  \addtocounter{footnote}{-1}%
  \endgroup
}

\begin{document}
\title{Domain Adaptation of LLMs for Process Data}

\author{Rafael Seidi Oyamada\inst{1} \and
Jari Peeperkorn\inst{1} \and
Jochen De Weerdt\inst{1} \and
Johannes De Smedt\inst{1}
}
\authorrunning{Oyamada et al.}
\institute{
  Research Centre for Information Systems Engineering, KU Leuven, Belgium\\
  \email{\{firstname.lastname\}@kuleuven.be}
}

\maketitle              % typeset the header of the contribution
\begin{abstract}
In recent years, Large Language Models (LLMs) have emerged as a prominent area of interest across various research domains, including Process Mining (PM).
Current applications in PM have predominantly centered on prompt engineering strategies or the transformation of event logs into narrative-style datasets, thereby exploiting the semantic capabilities of LLMs to address diverse tasks.
In contrast, this study investigates the direct adaptation of pretrained LLMs to process data without natural language reformulation, motivated by the fact that these models excel in generating sequences of tokens, similar to the objective in PM.
More specifically, we focus on parameter-efficient fine-tuning techniques to mitigate the computational overhead typically associated with such models.
Our experimental setup focuses on Predictive Process Monitoring (PPM), and considers both single- and multi-task predictions.
The results demonstrate a potential improvement in predictive performance over state-of-the-art recurrent neural network (RNN) approaches and recent narrative-style-based solutions, particularly in the multi-task setting.
Additionally, our fine-tuned models exhibit faster convergence and require significantly less hyperparameter optimization.
\blfootnote{This work was supported in part by the Research Foundation Flanders (FWO) under Project 1294325N as well as grant number G039923N, and Internal Funds KU Leuven under grant number C14/23/031.}

\end{abstract}

\section{Introduction}

LLMs have gained significant traction across various domains in recent years, including PM. 
In particular, PPM (the branch of PM concerned with predicting future process states and the behavior of cases) has begun to leverage its capabilities to reason over process data, given natural language instructions for contextualization.
The rapid progress in artificial intelligence and the advances of these powerful models offer new opportunities to improve PPM accuracy and flexibility beyond what traditional predictive models have achieved.
% They form a natural fit for this purpose as they generate tokens they learned from a vast number of sequences, which were formed by a particular (textual) grammar and vocabulary, similarly to how process executions are enveloped.

Recent studies on LLMs for PM tasks mainly follow two directions. 
One uses prompt engineering to enable pretrained LLMs to interpret event logs through crafted instructions~\cite{KubrakBMND24prompt,Berti24pmllmbenchmark}, thereby transforming process data into text for querying the model for insights. 
The other reformulates logs into narrative-style datasets for direct use or fine-tuning~\cite{PasquadibisceglieAM24,RebmannSGA24}. 
Both rely on the models' general language skills, treating logs as plain text and overlooking domain adaptation techniques to process-specific sequences.
Therefore, in this paper, we systematically evaluate how (small) LLMs can be fine-tuned for multi-task PPM, using parameter-efficient fine-tuning (PEFT) to lower the training costs. 
We test different PEFT methods to assess their ability to handle structured process data and compare their predictive performance.
We adapt the LLMs to directly consume structured process data by replacing the language-based tokenization with a task-specific tokenization of activity labels (i.e., replacing the embedding layers). 
This modification enables LLMs to learn from process data in its native format, bypassing the need for natural language conversion, which allows us to evaluate their intrinsic capability to interpret sequential process information.
Our experiments take into consideration real-world event logs and compare pretrained models with RNNs and prompt-based approaches. 
We focus on two PPM tasks: next activity (NA) and remaining time (RT) prediction, where we evaluate the predictive performance under both single-task (separate models) and multi-task (joint model) setups.

Our results demonstrate that LLMs can converge within a few training epochs and require minimal hyperparameter optimization, making them more straightforward to work with than many current PPM solutions, which often demand extensive hyperparameter tuning (RNNs) or longer runtime (narrative-style fine-tuning).
Moreover, using these models for multi-task learning proves to be more robust and consistent, particularly due to notable improvements in RT prediction.
Furthermore, these findings highlight not only the potential of employing them for PPM but also their inherent capacity to interpret sequential structures, even when detached from their original natural language training context. 
%This potentially also opens the door to a broader set of use cases for LLMs in process mining, while also making a case to train process-specific foundation models.

The remainder of the paper is structured as follows:
We start by introducing key concepts and related works in~\autoref{sec:background}.
Subsequently, we motivate our work in~\autoref{sec:motivation} and introduce our proposal in~\autoref{sec:methodology}.
Finally, we present our results in~\autoref{sec:experiments} and conclude in~\autoref{sec:conclusion}.

\section{Background and Related Work}\label{sec:background}

\subsection{PM and PPM}\label{sec:background-pm}

In PM, an event log $L$ consists of a set of cases $P$ (a.k.a., process instances or process executions). 
Each case $p_c \in P$ comprises a trace $t_c$ (a finite, ordered sequence of events) and a case identifier $c$. 
We denote a trace as $t_c = \langle e_0, e_1, \dots, e_n \rangle$, where each event $e_{c,i}$ corresponds to the $i$-th recorded activity execution in the case. 
For simplicity, we omit the case identifier and write a generic event $e_i = (a, time)$, where $a \in \mathcal{A}$ is the activity label from the activity set $\mathcal{A}$ present in $L$, and $time$ is the timestamp of execution.
Alternatively, events can be composed of event or case attributes as well, but these are not considered in this paper since we focus on features common to all event logs, aiming for a more robust systematic evaluation.
%because our goal is to systematically evaluate different event logs by considering only features common to all of them.

PPM is a subfield of PM focused on forecasting future behavior of ongoing process instances using predictive models~\cite{RamaManeiroVL23}. 
Generally, three main types can be distinguished: predicting the next activity, estimating the remaining time, and predicting the outcome. 
Suffix prediction is also a common task, and it is often seen as an extended form of next activity prediction~\cite{TaxVRD17}.
This paper focuses on NA and RT prediction.
%, excluding outcome and suffix prediction tasks.
NA prediction focuses on the activity that will be executed at time step $i+1$, given an ongoing process at any time step $i \geq 0$.
It can be defined as the function $f_\text{NA}: t_i \to \mathcal{A}$, where $t_i$ represents the trace state at time step $i$, and $\mathcal{A}$ is the set of possible activities.
RT prediction forecasts the time difference between the current event timestamp and the last event timestamp, and can be defined as $f_\text{RT}: t_i \to \mathbb{R}^+$, for $\mathbb{R}^+$ being the output of a positive real number representing the estimated remaining time.
These tasks can be tackled individually (single-task learning) or jointly (multi-task learning). 
In the multi-task setup, one model is trained to predict both the next activity and the remaining time~\cite{TaxVRD17}, and can be defined as $f_{NA,RT}: t_i \to \langle \mathcal{A}, \mathbb{R}^+ \rangle$.
% In this paper, we use deep learning (DL) models to design both predictive function setups.

To train such models, event logs are typically converted into structured datasets by extracting prefixes (partial cases up to a timestamp $i$) to build a training set $D_p$. 
Alternatively, the usage of full traces $D_t$ in a many-to-many setup with teacher forcing has been recently proposed~\cite{RoiderZE24}, where the model receives the true previous event at each step to predict the next, avoiding error accumulation. 
For instance, given $t = \langle a, b, c, d, e \rangle$, the input $x_t = \langle a, b, c, d, e \rangle$ is paired with target $y_t = \langle b, c, d, e, \texttt{<eos>} \rangle$, where \texttt{<eos>} marks the end of the sequence.

\subsection{Applications of Large Language Models in PM}

LLMs have gained significant traction in both research and industry, including PM. 
Their use generally falls into two categories: prompt engineering and fine-tuning.
Prompt engineering leverages the model's pretrained capabilities by crafting structured inputs without modifying the model's parameters. 
Techniques like few-shot learning and chain-of-thought prompting improve performance by adding examples or logical reasoning steps within the prompt~\cite{Petrov0B24promptlim}.
Fine-tuning, in contrast, adapts the model by modifying its parameters, either by updating parts of the existing architecture or inserting lightweight adapter layers.
Approaches like freezing most of the model while training a small subset of layers~\cite{LuijkenKM23transferlr} or using low-rank adapters~\cite{HuSWALWWC22} fall under the PEFT paradigm, allowing models to adapt to new tasks with minimal overhead.
% Recent studies have explored both approaches in the context of PM. 
In PM, prompt-based methods use LLMs directly on event log data to perform tasks such as process model discovery, waiting time cause detection, and explain recommendations in prescriptive systems~\cite{Berti24pmllmbenchmark,BertiKHL024,LashkevichMAD24promptwaittime,KubrakBMND24prompt}.
These methods generally treat event logs as natural language inputs and rely on the LLM's semantic understanding.
For instance, transforming event logs into text leads to solutions for suffix prediction~\cite{PasquadibisceglieAM24}, next activity prediction and anomaly detection~\cite{RebmannSGA24}, and next event prediction based on the traditional XES format.
Finally, rather than relying on a natural language interface to LLMs,~\cite{LuijkenKM23transferlr} explores transfer learning from a small (4-layer) transformer model trained on a real event log and fine-tuned on other ones.
Their work provides useful insights, including the need for retraining embedding layers to adapt the domain-specific vocabulary and syntactic rules to new event logs.
\section{Motivation}\label{sec:motivation} % Overreliance on textual capabilities for process data

The works introduced in the previous section mainly leverage the textual and semantic understanding of LLMs for PM tasks, such as handcrafted prompts or narrative-style fine-tuning~\cite{PasquadibisceglieAM24,Berti24pmllmbenchmark}.
%More specifically, most research has focused on handcrafted prompts or narrative-style fine-tuning~\cite{PasquadibisceglieAM24,Berti24pmllmbenchmark}.
For instance, a recent evaluation targeting the semantic awareness of LLMs has been proposed~\cite{RebmannSGA24}, where ongoing processes are turned into narrative stories. % by listing the current activity sequences (prefix) and the possible next activities.
This approach, although insightful, has a small methodological flaw, since the proposed textual representation provides a list of possible next activities to be predicted, but this list includes only the activities possible for the ongoing variant instead of the complete list from the event log. Hence, this renders an unrealistic scenario for online settings since the complete process variant of an ongoing process is unknown.
% latter is limited to those seen in that specific control-flow variant, rendering it unrealistic for online settings, where the complete process variant (and hence the narrowed set of possible activities) of an ongoing process is unknown.
%where ongoing processes are turned into stories by explicitly stating the current input sequence of activities along with the possible next activities for that specific variant. 
%The goal is to evaluate if LLMs are capable of achieving PM goals based on the semantical capabilities of understanding sequences of activity labels.
%While this approach is insightful, it limits the set of possible next activities to those seen in the specific variant, rather than using the full set of activities from the training data. 
%However, this is unrealistic since it does not hold in online settings, where the final variant (hence the narrowed set of possible activities) of an ongoing process is unknown.
Next to this, relying solely on the semantic meaning of activity labels disregards other issues regarding their quality, arbitrariness, or language variation~\footnote{The BPI 2012 event log, for instance, has labels recorded in both English and Dutch.}, and how these issues could affect results.
%On the other hand, all these works also disregard other issues like the quality or the language they are recorded

In contrast, our goal is to bring a systematic approach to leverage the capabilities of such models for PPM tasks via fine-tuning.
The current textual abstraction, while convenient, raises a fundamental misalignment: event logs are not linguistic artifacts that adhere to natural language grammar and vocabulary underpinning the syntax and semantics, but rather use a smaller alphabet (set of activity labels) with a more structured syntax according to domain-specific behavioral relations.
Suppose prompts are crafted using textual representations of event logs. 
In that case, the model might leverage prior semantic information that is not necessarily contextualized for process data compared to an adapted, fine-tuned model. 
The latter solution will force a pretrained model to adapt its knowledge to a new domain, through aspects such as (re)training small sets of weights, positional encoding, and process-specific embedding layers, which can better reflect the structure and behavioral relations of the process

Furthermore, prompt engineering requires expert knowledge and adds another dimension of complexity to carefully designing such instructions, which can be error-prone due to user mistakes (e.g., typos, bias, accidentally mixing up terms, etc.) or to the model's sensitivity (e.g., tiny changes in phrasing may lead to completely different results).
%Taking that into consideration, we claim that designing prompts is difficult because it depends on these factors, is prone to human error, and requires expertise to describe processes, define inputs, and specify the expected output format.
Recent studies show that prompt design strongly influences model behavior: small changes such as paraphrasing~\cite{Sclar0TS24promptlim} or reordering few-shot examples~\cite{LuBM0S22fewshotorder}, can significantly affect outcomes~\cite{Petrov0B24promptlim,MAHOWALD2024517}.
%In a nutshell, the literature has evidence that LLMs are very sensitive to small changes in prompt formatting:
%the outcomes can be significantly affected by prompt paraphrasing~\cite{Sclar0TS24promptlim} or by the order of examples provided in a few-shot setting~\cite{LuBM0S22fewshotorder}.
These issues highlight the fragility of prompt-based approaches, being prone to human error, and requiring expertise to describe processes, define inputs, and specify the expected output format. 
We further argue that semantic reasoning alone, as captured by LLMs trained on natural text, is insufficient for modeling structured process behavior.
%Thus, we argue that relying on the natural language-based reasoning capabilities of LLMs alone does not explicitly reflect the structure of real-world process semantics and syntax since they were originally trained to capture the meaning of words in a typical textual context.
Additionally, the domain adaptation via PEFT from modern LLMs to process data is an overlooked task in the current PPM literature.
Therefore, in the following, we present our proposal to address this gap.
\section{Methodology}\label{sec:methodology}

%To overcome the downsides of prompt engineering's extra layer of complexity and to leverage a more direct use of LLMs' contextual understanding of process behaviors independent of the semantic meaning of activity labels, we propose a systematic evaluation of modern fine-tuning techniques on open-source LLMs.
Based on the motivation stated above, we employ different PEFT strategies for different LLMs and validate their effectiveness by comparing them with state-of-the-art RNN-based and the recent prompt-based solutions, both in single-task and multi-task training settings.
The underlying motivation is that a single, versatile model capable of handling multiple prediction tasks is preferable to maintaining separate models, as it reduces deployment complexity and resource overhead.
Moreover, we argue that fine-tuning on process data \textit{as-is} should be more effective than fine-tuning on process data represented as mere text.
Our end-to-end evaluation framework is inspired by~\cite{KetykoMHD22dlframework} and can be structured into four main components: input layers, backbone, output layers, and the PEFT of these components.

\textit{Input layers (I).}
Consider the input event features of shape $(B, L, F)$, where $B$ is the batch size, $L$ is the sequence length, and $F$ is the number of features. 
The input layer turns these raw features into a common vector space.
For each categorical feature with vocabulary size $V_{cf}$ (i.e., the number of unique labels), an embedding layer maps each category to a dense tensor of size $E_{cf}$. 
On the other hand, $n$ numerical features can be straightforwardly passed through a linear projection layer to dimension $E_{nf}$. 
The outputs are then fused (usually by summation or concatenation) into a final embedding of size $E = E_{cf} \oplus E_{nf}$, where $\oplus$ is the employed fusion operation.
This produces an embedded input tensor of shape $(B, L, E)$.

\textit{Backbone (BB)}.
The backbone transforms this tensor into an output tensor $(B, L, D)$. 
It is composed of repeating blocks, each containing one or more layers. 
Let us consider a simple example: in recurrent networks, a block is a single recurrent layer. 
Thus, stacking $n$ blocks creates an $n$-layer deep RNN. 
Similarly, in decoder-only transformers, a block comprises a multi-head self-attention layer, a feed-forward layer, residual connections, and layer normalization.
% The depth perspective in deep learning is derived from the idea of stacking more blocks in this component.
% Deeper backbones require careful setting of learning rates, normalization strategies, and regularization to ensure stable training and prevent overfitting. 
This block-based design makes it easy to compare models and fine-tuning techniques by varying models' sizes while keeping the input/output interface fixed.

\textit{Output layers (O)}.
These layers map backbone outputs to task-specific predictions. 
Similarly to the other components, design choices are considered, and one may attach one linear head per task, or insert a shared multi-layer perceptron network before branching into separate linear heads. 
For example, consider the problem of predicting the NA and the RT simultaneously.
Starting from the backbone output $(B, L, D)$, two linear heads can be implemented to run in parallel: one maps $(B, L, D) \to (B, L, A)$ for NA prediction (for $A$ classes), and the other maps $(B, L, D) \to (B, L, 1)$ for RT prediction. 

% When training networks from scratch, this setup is straightforward.
% However, when fine-tuning LLMs, it is important to highlight that in this paper, $I$ and $O$ from the pretrained architecture are disregarded, and new (and much smaller) layers are designed and trained from scratch.
% This is due to the fact that our goal is to leverage the pretrained capabilities of handling sequential behavior of the backbone for process data.
% This is accomplished by the input layers, especially the embedding layer of activities, which will learn the new activity label's meaning within this context.
% Upon this setup, we shall now describe how PEFT methods can be included.

\textit{PEFT of LLMs for Process Data}.
Finally, PEFT refers to training only a small part of the parameters rather than the full (large) model.
It aims to adapt such models to new tasks efficiently, avoiding the cost of full fine-tuning, which is often impractical for models with millions or billions of parameters.
In this work, we train new, smaller I/O layers from scratch to reflect process-specific features, while the backbone is either frozen or slightly modified.
Thus, we formally define this module as follows.
Let $P$ denote the full set of backbone parameter weights. We partition these into
$P = P_{fr}\,\cup\,\theta,\quad P_{fr}\cap\theta=\varnothing$,
where $P_{fr}$ are frozen parameters and $\theta$ are the trainable parameters. 
In this paper, we consider three PEFT settings (although the proposed approach is not limited to).
\textbf{Full freezing}: $\theta = \varnothing,\quad P_{fr} = P$, where all backbone parameters remain frozen and are not updated, i.e., the gradient calculation is disabled;
\textbf{Partial freezing}: $\theta \subset P,\quad P_{fr}=P\setminus\theta$, where only a selected subset of the backbone weights is selected to be updated; and \textbf{Adapter-based tuning}: $\theta = \phi,\quad P_{fr}=P$, where $\phi$ are new parameters added between layers of the frozen backbone.
The adapter-based tuning employed in this work is the recent low-rank adapter (LoRA)~\cite{HuSWALWWC22}.
Given an arbitrary layer from a backbone containing a weight matrix $W\in\mathbb{R}^{m\times n}$, LoRA introduces two low-rank matrices $B \in \mathbb{R}^{m\times r}$ and $A\in\mathbb{R}^{r\times n}$,
and replaces $W$ with $W' = W + B\,A$; thus, only $A$ and $B$ are learned, which reduces memory usage. %adding $r(m+n)$ parameters per layer, while keeping each layer's input-output shape unchanged. 
% This design freezes the backbone parameters $P$ and trains only $\phi=\{A,B\}$, greatly reducing both memory usage and computation time when adapting to new tasks.
LoRA has been applied to PPM~\cite{RebmannSGA24}, but only for fine-tuning narrative-style datasets, unlike our focus on adapting to process semantics. 
Additionally, while prior work often fine-tunes only the backbone, using the process data directly requires training the I/O layers from scratch to learn a smaller, domain-specific vocabulary of categorical event features.

\section{Experiments}\label{sec:experiments}

% In this section, we describe our experimental setup, define and answer our research questions, and discuss the advantages and limitations brought up by fine-tuned LLMs in multi-task PPM.

\subsection{Experimental Setup}

% In this section, we describe the employed event logs, the preprocessing steps, the design choices made for input and output layers, the backbones employed, the PEFT methods, and the hyperparameter search space for optimization.
% The entire codebase is publicly available in our repository~\footnote{\url{https://github.com/raseidi/llm-peft-ppm}}.

\textbf{Event Logs, Preprocessing, and Tasks}.
We use five well-known, publicly available event logs: 
BPI12,\footnote{\url{https://data.4tu.nl/articles/dataset/BPI\_Challenge\_2012/12689204/1}}, 
BPI17,\footnote{\url{https://data.4tu.nl/articles/dataset/BPI_Challenge_2017/12696884/1}}, 
and three versions of BPI20:\footnote{\url{https://data.4tu.nl/collections/_/5065541/1}} 
Request for Payment (BPI20RfP), Prepaid Travel Costs (BPI20PTC), and Permit Data (BPI20PD).
These logs were chosen for their diversity in size, structure, and event complexity, as summarized in~\autoref{tab:log-props}.
% BPI17 is the largest in terms of trace length and the number of events. 
% BPI12 is also large, but it shows a high variation in trace lengths. 
% BPI20TPD has a higher number of unique activities. 
% Finally, BPI20PTC is relatively small, with medium-sized trace lengths, whereas BPI20PD is similar, but it contains very short traces.
%, including: accumulated time over the trace (starting from 0 on the first event), day of the month, day of the week (0-6), day of the year (0-364), hour of the day (0-23), minute of the hour (0-59), month of the year (0-11), second of the minute (0-59), seconds within the day (0-86399), and week of the year (0-51). 
During preprocessing, we discard cases with fewer than two events and extract time features from timestamps~\cite{OyamadaTJC24}.
Numerical features are z-score normalized; the only categorical feature is the activity label, indexed for embeddings.  We disregard alternative features in order to systematically evaluate using only common features for all datasets.
We use the unbiased split~\cite{WeytjensW21a} to divide data into train and test sets.
Event sequences are encoded using trace encoding~\cite{RoiderZE24}, instead of prefix encoding (see \autoref{sec:background}).
These sequences are used to train single-task and multi-task baselines, except for the prompt-based one, which only supports the former.
Due to the LLM cost, we only train them in the multi-task setup, as deploying multiple models is impractical.
Note that, for now, suffix prediction is omitted due to the fact that we are using decoder-only LLMs, so it raises the disalignment between input and output data: in our approach, the former consists of activity labels and time-related features, whereas the latter is composed of activity labels and the remaining time.

% \begin{table}[h]
\begin{wraptable}[9]{r}{6.3cm} % fixing these tables must be the last thing to do
\centering
% \scriptsize
\caption{Properties of event logs.}
\label{tab:log-props}
\resizebox{0.48\columnwidth}{!}{%
\begin{tabular}{@{}ccccc@{}}
    \toprule
    \textbf{Log} & \textbf{\# cases} & \textbf{\# evt.} & \textbf{\# act.} & \textbf{Trace length} \\ \midrule
    BPI20PTC     & 2099              & 18246            & 29               & 8.6927±2.3            \\
    BPI20RfP     & 6886              & 36796            & 19               & 5.3436±1.5            \\
    BPI20TPD     & 7065              & 86581            & 51               & 12.2549±5.6           \\
    BPI12        & 13087             & 262200           & 24               & 20.0351±19.9          \\
    BPI17        & 31509             & 1202267          & 26               & 38.1563±16.7          \\ \bottomrule
    \end{tabular}%
    }
% \end{table}
\end{wraptable}

\textbf{Input and Output Layers}.
To ensure comparability and for ablation purposes, all models use identical I/O-layers (i.e., all backbones, whether trained from scratch or fine-tuned). 
The input layer comprises an embedding layer for categorical data and a linear projection for numerical features, which are summed and result in a latent representation of dimension $n_embed$. 
The output layer consists of a single linear head per target task: two heads in multi-task settings (NA and RT), and one in single-task setups.
% In the suffix prediction setup, we use the CRTPD module from~\cite{GunnarssonBW23} to predict $n$ steps ahead given a prefix, where $n$ is defined as the same value for prefix length.

% heldepsk ref: \footnote{\url{https://doi.org/10.4121/uuid:0c60edf1-6f83-4e75-9367-4c63b3e9d5bb}}
\textbf{Backbones and PEFT Methods}.
The first baseline is a state-of-the-art recurrent model using the well-known Long Short-Term Memory (LSTM)~\cite{RamaManeiroVL23}. 
As a second baseline, we adopt the best transfer learning setup from~\cite{LuijkenKM23transferlr}, which trains a vanilla transformer on the Helpdesk event log and fine-tunes it on others. 
% We pretrain this model on the Helpdesk log, where it showed the best reported results. 
Since the vanilla transformer is outdated and small, we replace it with GPT-2 (named PM-GPT2 by us) for fair comparisons, the latest open GPT model with about 0.1 billion parameters. 
As a third baseline, we extend the narrative-style S-NAP model~\cite{RebmannSGA24} to our datasets, which transforms traces into prompts and fine-tunes a Llama model using LoRA (the original architecture is kept unchanged since it makes use of its textual capabilities). 
However, the original version writes prompts by listing the possible activities to be predicted for the ongoing trace variant only, which is unrealistic since the model cannot see which variant will result from an ongoing execution.
To ensure fair comparison, we adapt their narrative-style input to use all the activities from the train set instead. 
Finally, in line with open science and because fine-tuning requires model access, we do not include up-to-date large models for in-context learning and only use open-source models: Qwen2 (0.5B parameters) and Llama3.2 (1B parameters), to compare architecture and scale. Smaller models are also preferable for fast inference in production.

\textbf{Hyperparameter Search Space}.
We perform a grid search to evaluate model scaling, such as increasing the RNN size to compete with LLMs, as it allows systematic evaluation of fixed configurations without relying on sequential heuristics of automated methods, like Bayesian optimization.
% Grid search is preferred in this scenario to ensure reproducibility, transparency, and parallel execution, as it allows systematic evaluation of fixed configurations without relying on sequential heuristics of automated methods, like Bayesian optimization.
For LSTMs, we optimize the following hyperparameters: the number of layers (1 to 6), learning rates (5e-4, 1e-4, 5e-5), embedding (32, 128, 256, 512), hidden (128, 256, 512), and batch sizes (32, 64, 256). 
% For LLMs, we set the embedding size to the accepted size (768 for PM-GPT2, 896 for Qwen, and 2048 for Llama). 
For freezing, we choose full freezing or unfreeze specific sets of layers for fine-tuning: (0), (0,1), (-1), (-1, -2), where -1 and -2 refer to the last and the penultimate layers, respectively. 
For LoRA, we set $r$ to 256 and $alpha$ to $2r=512$ as suggested in~\cite{Biderman24loraft}. 
Our goal is to minimize fine-tuning effort while keeping computational costs manageable compared to RNNs.
In this regard, we set 10 epochs for fine-tuning LLMs and 25 for training the RNNs from scratch.
The cross-entropy loss and mean squared error (MSE) are employed as optimizers during training for NA and RT predictions, respectively. 
Moreover, the NA accuracy and the runtime for training and validation are also reported.
The entire codebase is publicly available in our repository~\footnote{\url{https://github.com/raseidi/llm-peft-ppm}}

\subsection{Results}

\begin{wraptable}[25]{r}{7.2cm}
    % \begin{table}[h]
    \centering
    \scriptsize
    \caption{Top NA accuracy and RT MSE per LLM with PEFT, plus total/trainable parameters (RNNs: 100\%) and runtime.}
    \label{tab:best-performances}
    \resizebox{0.57\columnwidth}{!}{%
        \begin{tabular}{@{}cccccc@{}}
            \toprule
            \textbf{Dataset} & \textbf{Backbone} & \textbf{NA Acc.} & \textbf{RT MSE} & \textbf{\begin{tabular}[c]{@{}c@{}}\# params\\ (\%trainable)\end{tabular}} & \textbf{\begin{tabular}[c]{@{}c@{}}Runtime\\ (hours)\end{tabular}} \\
            \midrule
            \multirow{5}{*}{BPI20PTC}
                & Llama3.2  & \textbf{0.8517} & 0.8978          & 1.2e+09 (6\%)         & 0.082 \\
                & PM-GPT2   & 0.7721          & 1.0361          & 1.2e+08(0.05\%)       & 0.008 \\
                & Qwen2.5   & \textbf{0.8517} & \textbf{0.8706} & 6.3e+08(22\%)         & 0.118 \\
                & S-NAP     & 0.2351          & -               & 1.3e+09(3\%)              & 0.349 \\
                & MT-RNN    & 0.6516          & 1.0484          & 8.9e+04               & 0.004 \\
                & ST-RNN    & 0.8052          & 0.9942          & 2.1e+05               & 0.003  \\
            \midrule
            \multirow{5}{*}{BPI20RfP}
                & Llama3.2  & 0.8253          & 0.6463          & 1.2e+09(0.01\%)      & 0.147 \\
                & PM-GPT2   & \textbf{0.8429} & \textbf{0.6366} & 1.2e+08(5\%)         & 0.002 \\
                & Qwen2.5   & 0.8362          & 0.6706          & 4.9e+08(0.02\%)      & 0.189 \\
                & S-NAP     & 0.4150          & -               & 1.3e+09(3\%)                 & 0.491 \\
                & MT-RNN    & 0.7707          & 0.6456          & 1.3e+05              & 0.006  \\
                & ST-RNN    & 0.8123          & 0.6446          & 8.6e+04              & 0.012    \\
            \midrule
            \multirow{5}{*}{BPI20TPD}
                & Llama3.2  & 0.7615          & \textbf{0.7663} & 1.2e+09(12\%)      & 0.597                \\
                & PM-GPT2   & 0.7454          & 0.9041          & 1.3e+08(7\%)       & 0.055                \\
                & Qwen2.5   & \textbf{0.7889} & 0.7729          & 6.3e+08(22\%)      & 0.250                 \\
                & S-NAP     & 0.4204          & -               & 1.3e+09(3\%)            & 2.892 \\
                & MT-RNN    & 0.7208          & 0.8865          & 9.3e+04             & 0.013                \\
                & ST-RNN    & 0.7603          & 0.8152          & 1.5e+05             & 0.013                     \\
            \midrule
            \multirow{5}{*}{BPI12}
                & Llama3.2  & 0.7454          & \textbf{0.8444} & 1.2e+09(12\%)         & 1.641                \\
                & PM-GPT2   & 0.7983          & 1.1405          & 1.3e+08(3\%)          & 0.135                \\
                & Qwen2.5   & 0.8162          & 1.0169          & 5.6e+08(12\%)         & 0.922                \\
                & S-NAP     & 0.1124          & -               & 1.3e+09(3\%)              & 6.033 \\
                & MT-RNN    & 0.7904          & 1.4058          & 1.6e+06               & 0.016                \\
                & ST-RNN    & \textbf{0.8358} & 1.4555          & 2.1e+05               & 0.011                     \\
            \midrule
            \multirow{5}{*}{BPI17}
                & Llama3.2  & 0.8687          & 0.5906          & 1.2e+09(6\%)           & 3.396                \\
                & PM-GPT2   & 0.8730          & \textbf{0.5683} & 1.3e+08(3\%)           & 0.559                \\
                & Qwen2.5   & 0.8637          & 0.5932          & 6.3e+08(22\%)          & 2.269                \\
                & S-NAP     & 0.1851          & -               & 1.3e+09(3\%)               & 34.99 \\
                & MT-RNN    & 0.8430          & 0.6613          & 8.9e+04               & 0.039                   \\
                & ST-RNN    & \textbf{0.8855} & 0.6862          & 1.4e+05               & 0.065                      \\
            \bottomrule
        \end{tabular}%
    }
    % \end{table}
\end{wraptable}

In this section, we discuss the results of our proposed methodology.
In order to guide the reported results, we define three research questions.
\textbf{RQ1}: Do LLMs fine-tuned for process data outperform existing methods, and what are the associated trade-offs?
\textbf{RQ2}: Can LLMs effectively be adapted to the process domain in order to learn multiple PPM tasks simultaneously?
\textbf{RQ3}: Which PEFT method works best for the addressed PPM tasks?

\textbf{Research question 1 (RQ1)}.
\autoref{tab:best-performances} shows, for each backbone and dataset, only the best scores, the total and trainable parameter counts, and the runtime for training and validating.
Three key insights can immediately be extracted from these results.
First, the MT-RNN setup never achieves the highest scores on any dataset, and S-NAP is significantly outperformed for all cases.
The issue with MT-RNNs lies in the fact that we are increasing the problem complexity by predicting multiple targets at once while preserving the architectural simplicity of recurrent networks.
The lower S-NAP results seem to indicate that the semantic capabilities of the employed LLM are not enough to learn process behaviors for real-world datasets, based on activity labels alone.
The low accuracies, as compared to the original work~\cite{RebmannSGA24}, are also explained by the prompt change to not include future information, as explained earlier (see~\autoref{sec:motivation}).
Interestingly, at first trial, S-NAP achieved an accuracy of 2\% for BPI12, and we noticed that was due to activity labels mixed in English and Dutch, which raises another limitation of narrative-style solutions if the employed LLM is not cross-lingual. 
After manually standardizing all labels to English and retraining the model, a significant improvement was achieved, as reported in the table.
Additionally, the ST-RNN remains a strong baseline for NA prediction: it significantly outperforms other models on BPI12 and outperforms, by a small margin, on BPI17. 
Regarding the RT prediction, it is competitive only on BPI20RfP and is outperformed on the other datasets.
Thus, although the ST-RNN excels on selected cases, it lacks consistency and significantly underperforms on NA for BPI20PTC and BPI20RfP.

Second, recurrent nets use orders of magnitude fewer parameters and less runtime than LLM-based solutions, highlighting the higher computational cost to improve predictive performance.
However, notice that the number of total parameters differs between the different RNN setups since they were optimized in a large hyperparameter search space (e.g., MT-RNN for BPI12 and BPI17).
Intuitively, this shows that although RNNs are smaller, they demand careful hyperparameter optimization.
Our LLMs, especially with LoRA and the employed default settings, require less tuning and clearly outperform RNNs and S-NAP in both single- and multi-task setups.
Still, ST-RNN in particular is a strong, low-cost baseline for NA but not for RT, while LLMs perform well on both.
S-NAP runs slower, mainly due to the narrative-style inputs that make sequences longer. 
% From this analysis, a valuable observation is that the capability of deploying one model for all tasks also simplifies maintenance.
% This is meaningful from the point of view that having only one model for all tasks requires less effort in terms of maintenance.
Lastly, among the three LLMs fine-tuned for process data, PM-GPT2 achieves the best RT performance on BPI17 and outperforms on both tasks on BPI20RfP.
However, it lacks consistency, underperforming on other datasets, unlike Llama and Qwen, which are very stable across all datasets. 
Thus, we can conclude that LLMs, when explicitly adapted to the process domain, can outperform current methods in both single- and multi-task setups at the cost of increasing computational cost (though it is still much lower than the narrative-style S-NAP) but decreasing the burden of hyperparameter optimization.

\begin{figure}[h]
  \centering
  \includegraphics[width=0.95\linewidth]{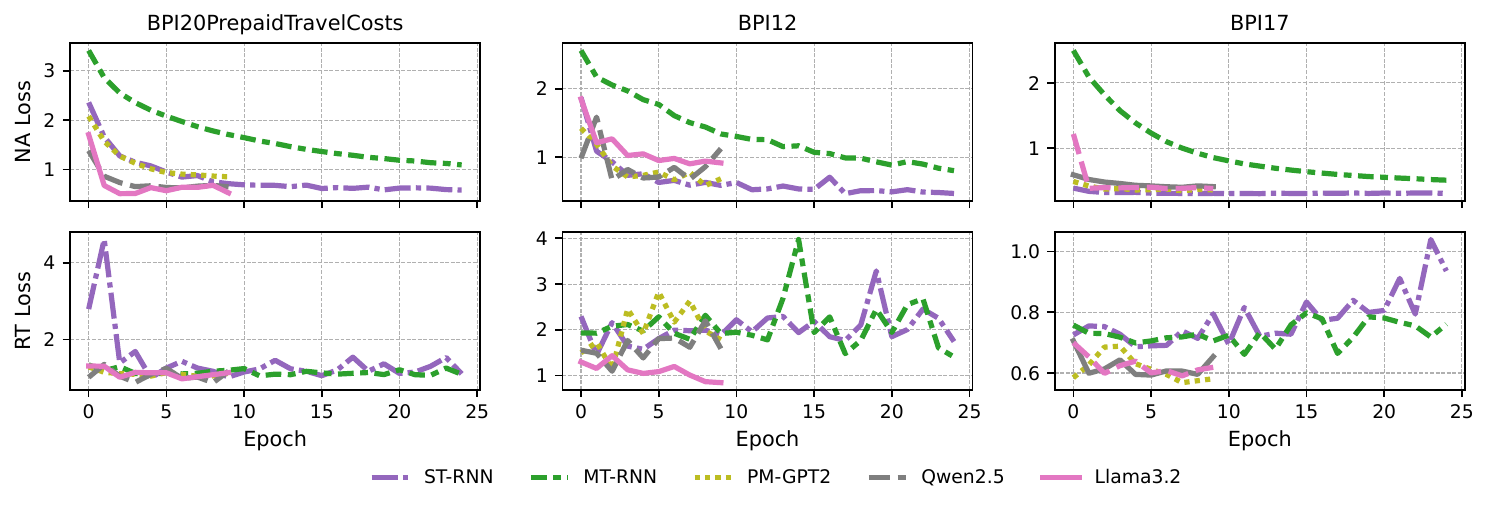}
  \caption{Curve losses for RNNs and LLMs fine-tuned for process data.}
  \label{fig:losses}
\end{figure}

\textbf{Research question 2 (RQ2)}.
We answer this question by assessing the loss curves of models on both NA and RT tasks, by reporting only the best models illustrated in~\autoref{fig:losses}. 
Due to the significantly lower accuracy, S-NAP is excluded.
We include only three datasets due to the lack of space, although the behaviors were similar for all the BPI20 datasets (remaining plots can be found in the online repository).
LLMs and ST-RNNs converge faster than MT-RNNs on NA prediction, with LLMs needing fewer than 5 epochs.
For RT prediction, LLMs constantly outperform single- and multi-task RNNs.
Moreover, these results also highlight the difficulties of RNNs on multi-task learning, as we can see on BPI17, where the MT-RNN converges on the NA task but underfits for RT.
It is noticeable how difficult the RT task is, regardless of the employed backbone, as most of them exhibit spiky behavior instead of a smooth curve.
Llama excels in RT prediction, being most consistent across datasets, whereas the other LLMs are not, showcasing that model size matters in this setup.
The particular spiky behavior presented by LLMs on the RT task is due to the fact that these models are originally trained as classifiers. 
Thus, adapting these smaller models to a regression task might require further efforts.
Nevertheless, in the multi-task setup, LLMs are capable of outperforming all the RNNs on the RT task and performing competitively to ST-RNNs on the NA task, which evidences their capabilities of learning both tasks at once.

\begin{figure}[h]
  \centering
  \begin{subfigure}[t]{0.4\columnwidth}
    \centering
    \includegraphics[width=\linewidth]{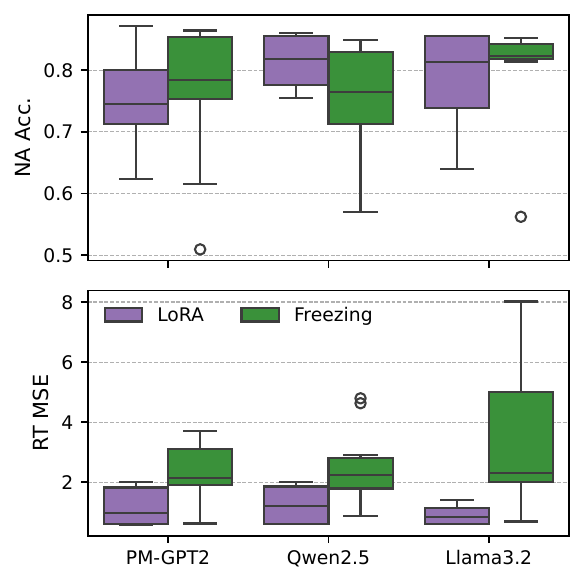}
    \caption{}
    \label{fig:loss-distribution}
  \end{subfigure}
  \hfill
  \begin{subfigure}[t]{0.5\columnwidth}
    \centering
    \includegraphics[width=\linewidth]{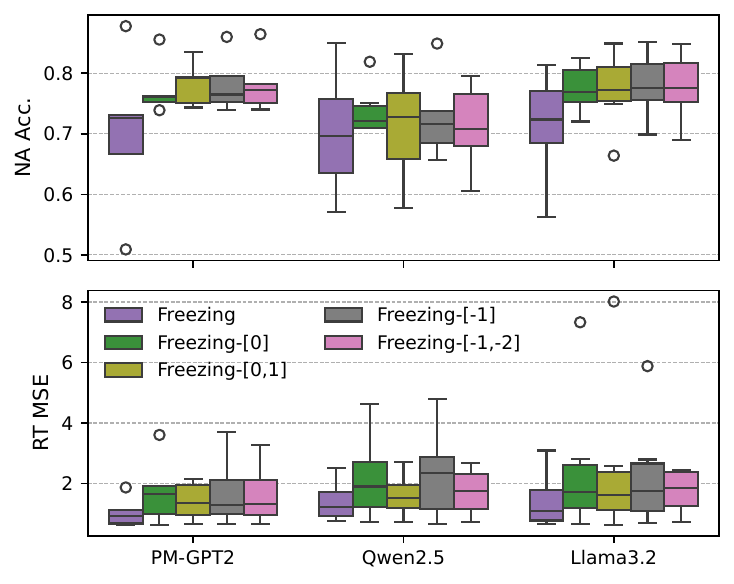}
    \caption{}
    \label{fig:loss-distribution-freezing}
  \end{subfigure}
  \caption{(a) Accuracy and MSE of each PEFT for each language model on all event logs used. (b) Accuracy and MSE for each freezing configuration on all event logs.}
\end{figure}

\textbf{Research question 3 (RQ3)}.
\autoref{fig:loss-distribution} shows the loss distribution for LoRA and Freezing configurations across all datasets. 
For the NA task, all PEFTs perform well: LoRA works better for Qwen, whereas freezing layers work better for Llama and PM-GPT2. However, for PM-GPT2, the optimal result is achieved by LoRA.
For RT, LoRA clearly outperforms freezing. This reinforces the previously introduced idea that LLMs, trained for classification (in the form of next token prediction), struggle with regression unless explicitly adapted.
LoRA overcomes this limitation by adding new layers that adapt the LLM to the regression task. This lowers the loss distribution for all models and emphasizes the fact that LLMs need to be properly adapted somehow, where the insertion of new adapter layers instead of simply fine-tuning whole layers is an effective alternative.
In~\autoref{fig:loss-distribution-freezing}, we provide a more detailed view by showing the loss distributions for each setup. 
For NA prediction, all LLMs require a few layers to be fine-tuned since freezing all the weights does not perform well.
Although for Qwen freezing all layers might achieve a high accuracy, it is very inconsistent compared with the other configurations, since it has the highest variation.
This evidence shows again that LLMs are next token classifiers; hence, retaining a few of their original layers and consequently adapting the model to the process domain, works better.
In contrast, fully freezing the model proves more effective for RT prediction.
Although this might seem counterintuitive regarding the previous insights for NA, it highlights again the fact that LLMs are not inherently suited for regression tasks, and the good performance is mainly driven by training only input and output layers.

\subsection{Distinctions, Limitations, and Future Directions}

This work introduces modern fine-tuning techniques to adapt LLMs to the process domain, aiming to reduce the reliance on prompt engineering or narrative-style reformulations
Instead, we use PEFT to explicitly teach process-specific patterns, going beyond general reasoning abilities.
A valuable insight from our experiments includes the ease with which these models present to perform next token classification, more in line with their original training objectives, whereas they might struggle with regression tasks.
Still, our approach outperforms existing solutions, based on both RNNs and prompt engineering, suggesting that targeted adaptations are effective and necessary to bridge this performance gap.
These results are much faster than narrative-style approaches, even when using open-source, lightweight models, which make them suitable for production. 
The speed-up comes from avoiding long prompt-based inputs, which slow down inference. 
Still, some limitations remain. While PEFT reduces cost, it requires more trainable parameters than RNNs; future work could explore quantization to shrink models further. 
Key tasks like process discovery and anomaly detection remain challenging due to their mismatch with standard training formats. Finally, we used LoRA with default settings, which could be tuned to reduce memory usage.

\section{Conclusion}\label{sec:conclusion}

This study systematically evaluates fine-tuning methods for adapting LLMs to predictive process monitoring. 
Unlike prior works that mainly focus on LLMs' abilities to understand language semantics, we focus on domain adaptation through fine-tuning to directly use process event data.
By going beyond prompts and narratives, our results show that fine-tuned LLMs can outperform traditional PPM models and narrative-style approaches in both single- and multi-task NA and RT prediction.
% By going beyond prompts and narratives, we show that LLMs can effectively model process dynamics, offering useful insights for improving PPM performance and reliability, and process mining as a whole.

%\section*{Acknowledgments}
%This work was supported in part by the Research Foundation Flanders (FWO) under Project 1294325N as well as grant number G039923N, and Internal Funds KU Leuven under grant number C14/23/031.

%Bibliography
\bibliographystyle{splncs04}
\bibliography{references}

\end{document}